\documentclass[letterpaper, 10 pt, conference]{ieeeconf}  
\usepackage{multirow}

\IEEEoverridecommandlockouts                              

\usepackage{graphics} 
\usepackage{epsfig} 
\usepackage{mathptmx} 
\usepackage{times} 
\usepackage{amsmath} 
\usepackage{amssymb}  
\usepackage{subfigure}
\usepackage{tabularx}
\usepackage{makecell}
\usepackage{cite}
\usepackage{stfloats}
\usepackage[font=footnotesize]{caption}

\title{\LARGE \bf
POA: Passable Obstacles Aware Path-planning Algorithm for Navigation of a Two-wheeled Robot in Highly Cluttered Environments
}

\author{Alexander Petrovsky$^{1}$, Yomna Youssef$^{1}$, Kirill Myasoedov$^{1}$, Artem Timoshenko$^{1}$, \\ Vladimir Guneavoi$^{1}$, Ivan Kalinov$^{1}$, and Dzmitry Tsetserukou$^{1}$
\thanks{$^{1}$All authors are with Skolkovo Institute of Science and Technology, Moscow 143026, Russia 
{\ ivan.kalinov@skolkovotech.ru \{aleksandr.petrovskii, yomna.youssef, artem.timoshenko,  kirill.myasoedov, vladimir.guneavoy, d.tsetserukou\} @skoltech.ru}}%
}

\begin{document}
\maketitle
\thispagestyle{empty}
\pagestyle{empty}


\begin{abstract}

This paper focuses on Passable Obstacles Aware (POA) planner - a novel navigation method for two-wheeled robots in a highly cluttered environment. The navigation algorithm detects and classifies objects to distinguish two types of obstacles - passable and unpassable. Our algorithm allows two-wheeled robots to find a path through passable obstacles. Such a solution helps the robot working in areas inaccessible to standard path planners and find optimal trajectories in scenarios with a high number of objects in the robot's vicinity. The POA planner can be embedded into other planning algorithms and enables them to build a path through obstacles. Our method decreases path length and the total travel time to the final destination up to 43\% and 39\%, respectively, comparing to standard path planners such as GVD, A*, and RRT*. 

\end{abstract}


\section{Introduction}
\subsection{Motivation}

Autonomous robots integrate into outdoor operations more deeply as different robotic solutions emerge that provide high passability in highly unstructured environments, \cite{sun2019underactuated}. Also, various navigation techniques appear that allow mobile and flying robots safely explore their environment \cite{cai2020mobile, kalinov2020warevision, karpyshev2022mucaslam, kalinov2019high, yatskin2017principles, kalinov2021impedance, protasov2021cnn, kalinov2021warevr}. One of the domains where high mobility and intelligent navigation systems play a crucial role in space robotics, i.e. Mars Exploration Rovers (MERs). MER is a wheeled motor vehicle that travels on the surface of planet Mars. The primary goal of these rovers is to explore the planet's territory. During the last 24 years there were six successful Mars exploration missions: Sojourner, Spirit, Curiosity, Opportunity, Tianwen-1, and Perseverance, \cite{farley2020mars}, \cite{zou2021scientific}.  

Over the past two decades, many MERs with increased mobility has proposed. The vast majority of concepts that focus on autonomous robots emphasize the increased mobility of planetary rovers. One of the last MER missions, Perseverance consists of a traditional six-wheeled mobile rover with Ingenuity Mars Helicopter. Such a combination allows Mars exploration missions to investigate previously inaccessible areas and increase overall mobility. One of the solutions that have high mobility and can freely overcome various obstacles is a two-wheeled mobile platform \cite{vidhyaprakash2019positioning}, \cite{deng2010two}. 

In our previous work \cite{petrovsky2022two}, we described the robot design in detail and discussed the possibility of using a swarm of two-wheeled robots, as highly mobile platforms, for Mars exploration. We introduced metrics for modular surface exploration systems assessment, such as mission lifetime, exploration speed, and mission cost. These metrics proved the utility of a two-wheeled swarm concept for outdoor exploration tasks. This article aims to solve the path planning problem for two-wheeled robots to optimally build a path in the outdoor environment concerning various surrounding obstacles.

\vspace{-0.5em}
\subsection{Literature review}

\begin{figure} [!t]
\begin{center}
\includegraphics[width=7.3cm]{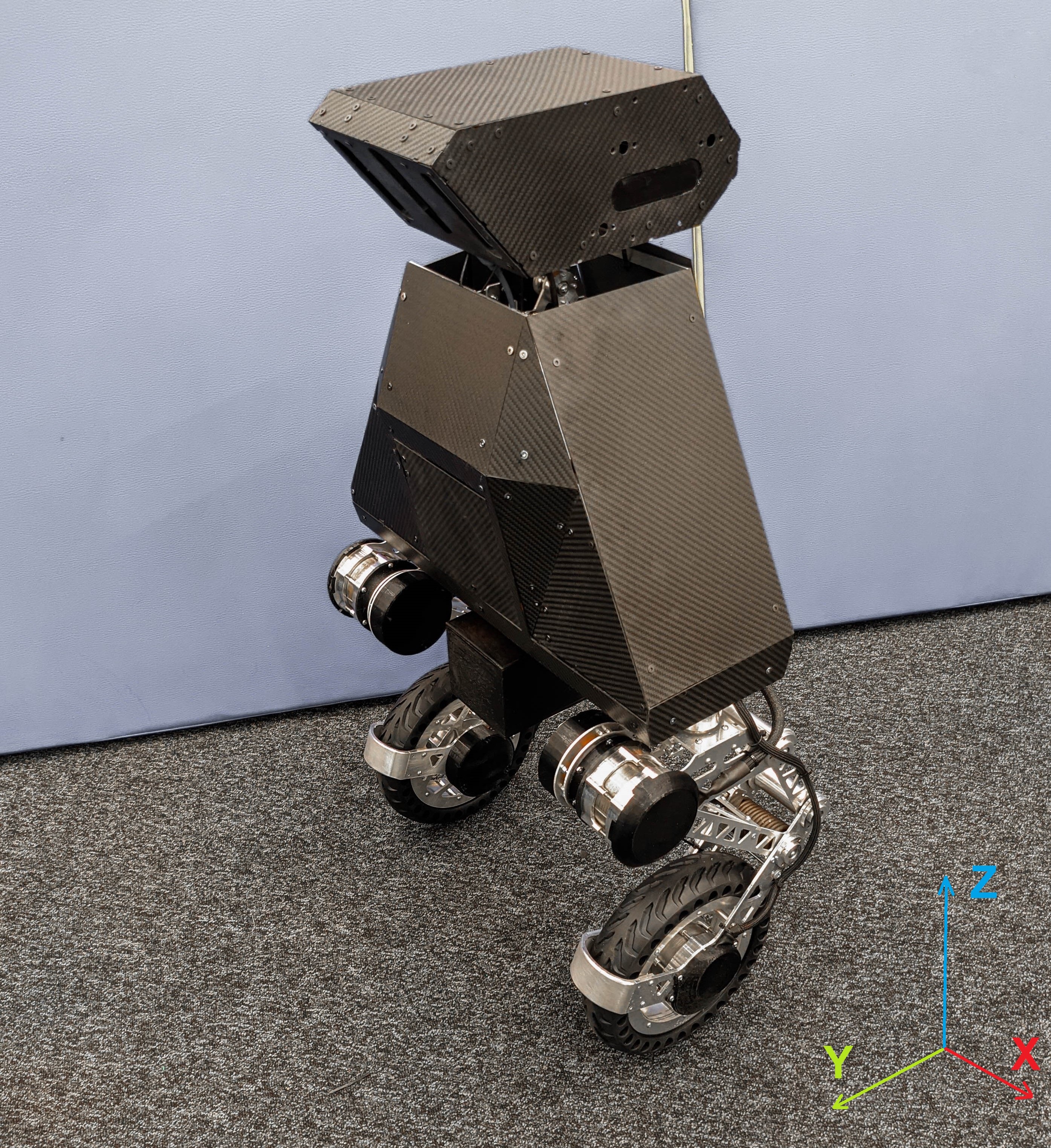}
\caption{The two-wheeled robot prototype}
\label{fig1}
\vspace{-2.5em}
\end{center}
\end{figure}

Since two-wheeled robots have higher ground clearance and better maneuverability, standard planning algorithms may not be as effective for them as for the other wheeled robots. That is why a two-wheeled robot needs to perform an object classification to distinguish passable obstacles or slight unevenness on the ground from non-passable obstacles that robot should avoid. Since we base our experiments on a relatively small two-wheeled robot (\ref{fig1}), using RGB-D cameras over 3D LiDARs is preferable. RGB-D cameras take up much less space than LiDARs - they are lighter and require less computational power for real-time 3D point cloud processing. Many solutions were proposed using RGB or D channels information to classify obstacles in an outdoor environment. 
Wang et al. \cite{wang2021vehicle} reached 89.3\% accuracy with 1 second of overall computation time per point cloud using per-object classification algorithm. In the paper by Zhang et al. \cite{zhang2012object}, authors use geometric and topological information and local neighbourhood-related features. Their work demonstrated classification accuracy higher than 97\% using SVM as objects classifier. 

Many works are concerned with path planning in scattered environments. They use different classification tools to distinguish between passable and unpassable terrain points and how this information is used in path planning. Menna et al. \cite{menna_gianni_ferri_pirri_2014} presented a novel framework for 3D autonomous navigation where the point clouds are segmented and classified into four categories: ground, walls, ramp or stairs, and surmountable obstacles. The robot presented in the article can elaborate a point cloud composed of 55000 points and generate a path of less than 3.5 s.
Raghavan et al. 

Only a few mentioned path planning algorithms utilise two-wheeled robots increased mobility. Moreover, none of the mentioned approaches allows building a path through small obstacles without changing robots configuration. There is a need for a solution that enables a two-wheeled robot to optimally build a path in the outdoor environment taking into account different classes of surrounding objects.

\subsection{Problem statement}

As we can see, there is a lack of path planning approaches for two-wheeled robots that can stably work in an unstructured environment. As the main idea of our research states, there is a need for a solution that enables a two-wheeled robot to optimally build a path in the outdoor environment concerning surrounding obstacles. Building a path around obstacles is not always the most optimal way of navigation for a two-wheeled robot. Our idea is that it is possible for TWR to find a path through some of the obstacles. In some cases, such a path can be the most optimal, in others - it can be the only path to reach a certain point. 

\textbf{Our main contributions to this work are:}

\begin{itemize}
    \item The POA planner allows two-wheeled robots to build a path through small (passable) obstacles decreasing traversal time and path length to reach the final point in 3D outdoor environment;
    \item The POA planner complements existing path planners, allowing them to safely build a path through some objects and improving their performance in highly cluttered environments;
    \item The POA planner classifies objects in 3D environment using only RGB-D cameras, allowing robots to distinguish passable objects and build a path through them.
\end{itemize}


%
%
%
%
%
%
%
%
%
%
%
%


\section{Methodology} 
\label{3}

\subsection{System overview} 
\label{over}

This section briefly explains the two-wheeled robot system architecture. 
The two-wheeled robot is designed to go through obstacles up to 28 cm high, keeping the vertical position. The robot consists of several modules: motor wheels, adaptive suspension for each leg (provides stabilization in $ZX$ plane (Fig. \ref{fig1})), robot body with computational systems and necessary electronics, and robot head with cameras and sensors for autonomous localization, navigation, and collision avoidance. The system architecture is presented in more detail in our previous work \cite{petrovsky2022two}. The robot uses two RealSense D435i RGB-D cameras and a RealSense T265 stereo camera to collect environmental data and send coordinate frames (TF), raw RGB images, and a 3D point cloud to the SLAM (Simultaneous Localization And Mapping) algorithm (RTAB-Map). For object classification, the robot uses the multi-branched ERFNet neural network that takes the data from RGB-D cameras. To keep the robot upright when moving, we use the LQG (Linear Quadratic Gaussian) controller \cite{patra2020design}. LQG receives input from IMU (Inertial Measurement Unit) and Inductive Angle Sensors and sends torque output to torque motors.

\subsection{Point cloud classification and mapping}

To solve a point cloud classification problem, we use a method of RGB image segmentation and its back projection on a 3D point cloud. We segment several images and project them on a 3D point cloud, where the final label of each point is chosen by the majority votes method. After the segmentation, the robot estimates stones' dimensions to classify them as \textit{passable} obstacles - small enough for the robot to pass them between its legs, and \textit{unpassable} - large objects that the robot should avoid. To do this, our system should be able to distinguish different stone instances, for example, via instance segmentation. Our system uses the approach of clustering loss function that enables real-time computations while keeping high accuracy. In the two-wheeled robot system, the point cloud classifier module simultaneously receives images, point cloud, and camera info from the D435i cameras. The system segments the images and outputs the mask for every instance of the stone class. It back-projects the masks on the point cloud. Then, it estimates the the stones dimensions to classify them as passable and unpassable. 

\begin{figure} [!b]
\vspace{-1em}
\begin{center}
\includegraphics[width=8.4cm]{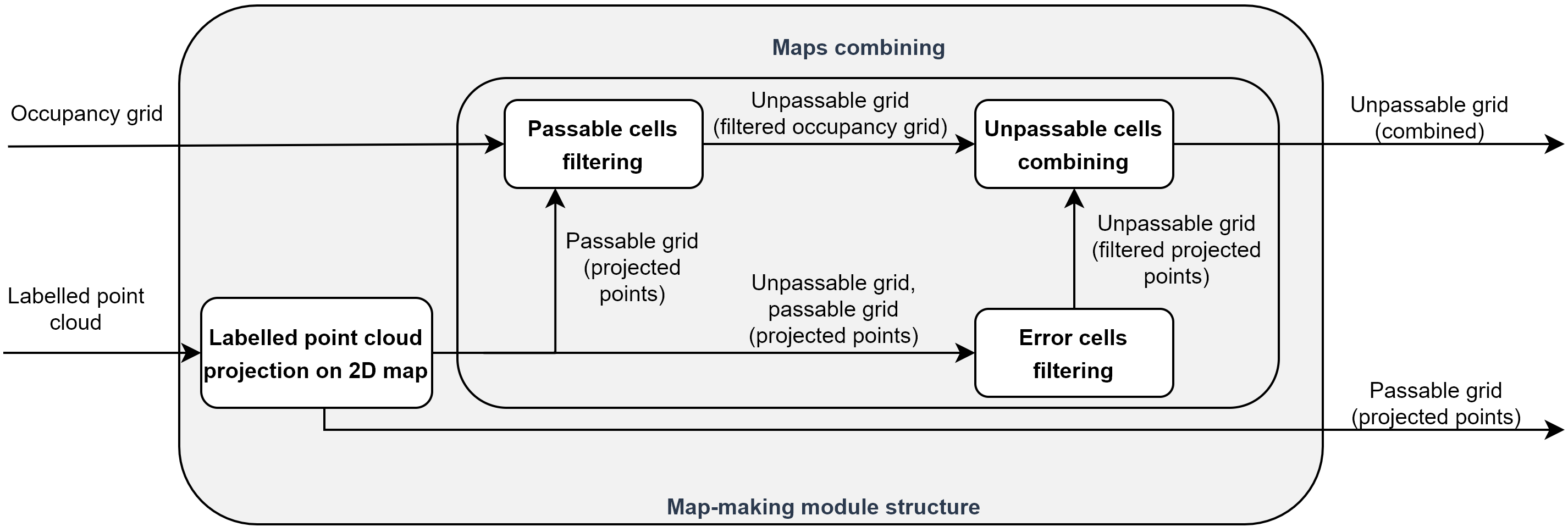}
\caption{The map-making module structure}
\label{map2}
\end{center}
\end{figure}

The labelled point cloud is sent to the map-making module. It projects the points onto the horizontal plane, creating two occupancy grids with cells representing passable and unpassable areas (passable grid and unpassable grid). Fig. \ref{map2} depicts the structure of the map-making module in detail. It receives the occupancy grid from the SLAM module and uses the passable grid as a mask to filter out the occupied cells corresponding to the passable stones. Also, the module saves all the point clouds from the previous messages to get the common labelled point cloud of investigated environment.

Our robot filters out camera odometry noise, to estimate 3D point cloud points' position and project them correctly . Then it creates a probability map that takes into account previous information about stones' position and finds the average result to adjust the outliers. Projecting the point cloud on the 2D grid the map-making module finds the probability for every cell that this is occupied, according to current measurements:
\begin{equation}
p_{current} = 0.5 + p \cdot n_{stone} - p \cdot n_{environment}
\end{equation}
Where $p_{current}$ is the probability of a cell occupation, according to the current measurements, $p=0.01$ is the parameter that adjusts the occupation probability, $n_{stone}$ is the number of projected stone points in the cell, $n_{environment}$ is the number of projected environment points in the cell. Using the acquired measurement model, the system calculates the log odds of the measurement probabilities and adds them to the previous log odds of the grid:

\begin{equation}
log \, odd_{xy}^{i} = log(\frac{p_{current}}{1 - p_{current}}) + log \, odd_{xy}^{i-1},
\end{equation}
while $log \, odd_{xy}^{0} = 0$. The final probability that the cell is occupied:
\begin{equation}
p_{final}^{i} = \frac{1}{1 + exp(-log \, odd_{xy}^{i})}
\end{equation}

To get rid of shifted stone points in the common point cloud, we filtered out the wrong stone points using the resulting occupancy grid as a mask.

 \subsection{2D POA Path planner}
The path planner module generates a collision-free path from the current robot pose to the defined goal pose. The module utilizes the labelled occupancy grid and point cloud to incorporate an awareness of passable obstacles into the planning pipeline. The operation pipeline can be described as follows: an initial trajectory is created on the unpassable occupancy grid, and then waypoints along the trajectory are checked to see if either wheel collides with a passable obstacle. Finally, in the case of a collision prediction at a waypoint, the path segment around this waypoint is updated to avoid collision with passable obstacles.

\begin{figure}[t!]
    \centering
    \vspace{0.5em}
    \includegraphics[width=8cm]{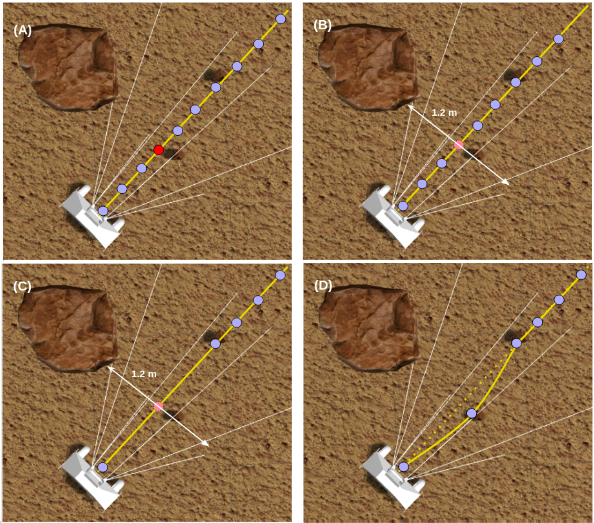}
    \caption{POA path update: (A) - A collision with a passable obstacle is detected at the red waypoint; (B) - Alternative waypoints are generated along the direction perpendicular to the robot's heading; (C) - $n_{clear}$ waypoints are cleared before and after the collision; (D) - Dubins curve trajectories are created to and from the closest and safe alternative waypoint.}
    \vspace{- 2em}
    \label{fig:POA path update}
\end{figure}

The initial trajectory is generated by a graph search algorithm Generalized Voronoi Diagram (GVD) based on Dynamic Voronoi Diagram on the unpassable occupancy grid. That way two-wheeled robot does not take into account passable obstacles.
At the next step the 2D POA planner evaluates the initial trajectory by checking if the robot's wheels collide with passable obstacles. This is accomplished by an elliptical collision zone that encompasses each wheel. The hyperparameter $n_{skip}$ is a constant value that defines the number of waypoints skipped before the next collision check. If a collision with a passable obstacle is detected at $waypoint_i$, this waypoint is declared hazardous. After that, an alternative $waypoint_i$ is generated by shifting the hazardous waypoint in the direction perpendicular to the robot's orientation. The value of the shift distance ranges from -0.6 m to 0.6 m with a step size of 0.05, resulting in a total of 24 alternative waypoints. POA then removes $n_{clear}$ waypoints before and after the hazardous waypoint. After that, the path segment from $waypoint_{i-n_{clear}}$ to an alternative $waypoint_{i}$ is updated with a Dubins curve trajectory\cite{Dubins1957OnCO}. The alternative $waypoint_{i}$ is chosen such that the corresponding Dubins curve trajectory is free from collisions with passable or unpassable obstacles, and the trajectory has the least deviation from the original path. To provide a smooth transition back to the original path, another Dubins curve trajectory is generated from $waypoint_{i}$ to $waypoint_{i+n_{clear}}$. This is further demonstrated in Fig. \ref{fig:POA path update}.
\vspace{- 0.1em}

\subsection{3D Path Planner}

In this section, we describe the extended POA path planner to find feasible 3D paths on non-planar terrain. A labelled point cloud map is used to represent non-planar terrain. The labelled unordered point cloud map is obtained by labelling the point cloud map from the SLAM algorithm. To do that, the robot uses the labelled occupancy grid maps after point cloud classification. The next steps for obtaining a feasible 3D path are as follows: the labelled point cloud map is preprocessed to remove any shifts or noise and downsampled to an appropriate density. After that, the 2D POA path (generated in the previous section) is projected onto the point cloud map. The 3D path's feasibility is then evaluated using the estimated roll angle and pitch angle of waypoints along the 3D trajectory. If one of these properties exceeds a predefined threshold, the 3D trajectory becomes infeasible, and the 2D path is updated. This process is repeated until a feasible 3D path is found. 

\begin{figure}[t!]
\begin{center}
\vspace{0.5em}
\subfigure[Gazebo environment]{
\resizebox*{8.4cm}{!}{\includegraphics{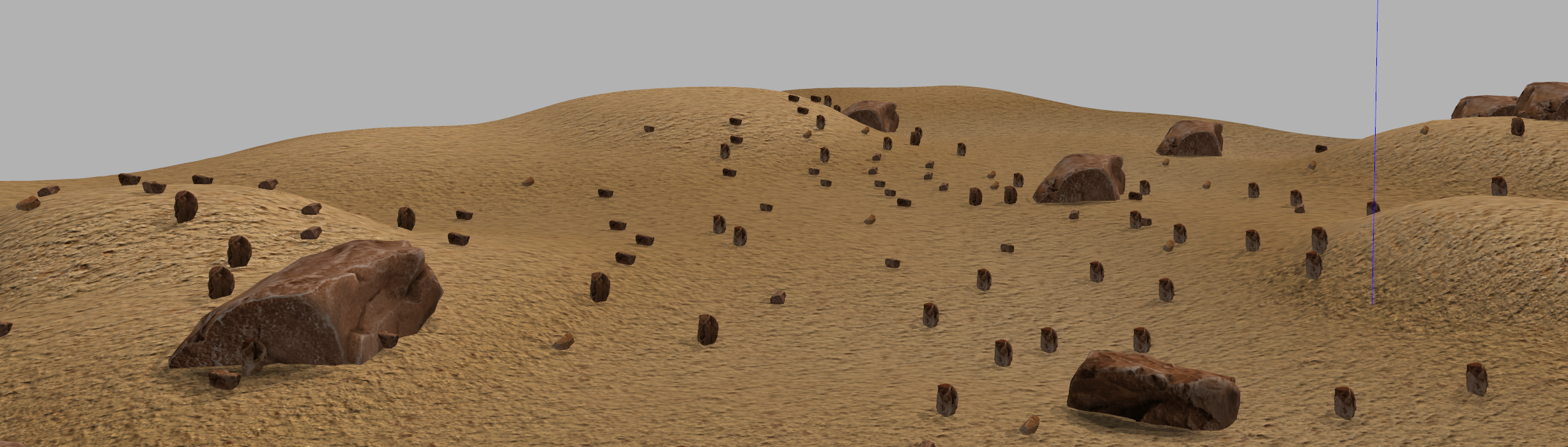}}}
\subfigure[Labeled point cloud after outlier removal and downsampling;]{
\resizebox*{4.1cm}{!}{\includegraphics{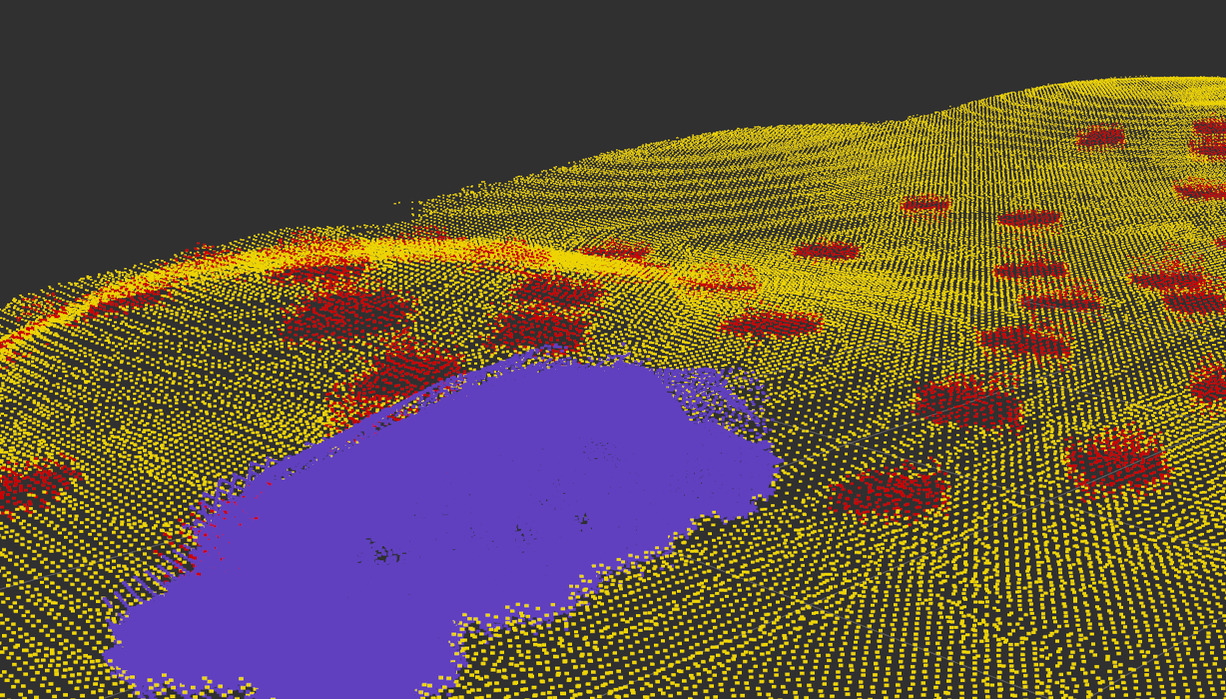}}}
\subfigure[Surface reconstruction of "free-space"]{
\resizebox*{4.1cm}{!}{\includegraphics{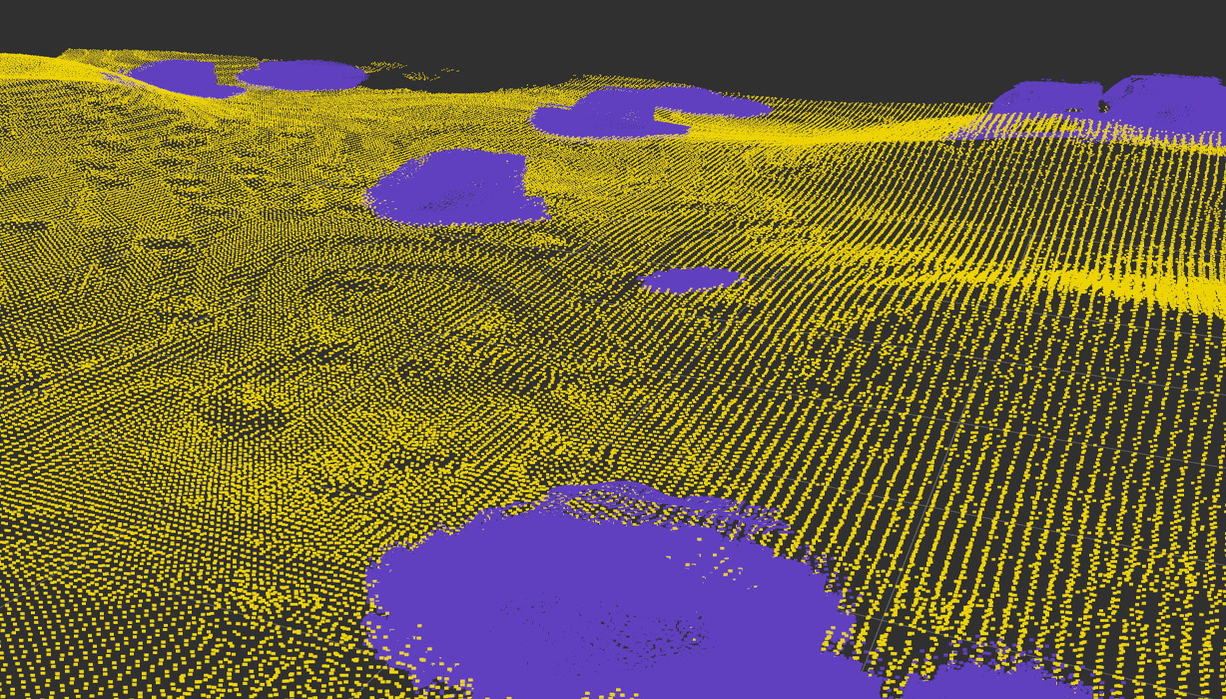}}}
\vspace{-0.5em}
\caption{Point cloud map preprocessing: "free-space" - yellow, "unpassable-obstacles" - purple, and "passable-obstacles" - red.}
\label{pc_pics_new}
\vspace{-2em}
\end{center}
\end{figure}

The preprocessing of the point cloud map, as demonstrated in Fig. \ref{pc_pics_new} (b), is initiated by performing obstacle inflation on the passable and unpassable occupancy grid maps. Then, the resulting grid maps are used to label the point cloud map. The labelled point cloud then undergoes outlier removal and downsampling using the Point cloud library in ROS pcl\_ros's\cite{rusu20113d}. Finally, we perform surface reconstruction on the "Free-space" point cloud using Radial basis function (RBF) interpolation. The reconstructed "Free-space" point cloud represents the terrain surface without any obstacles (Fig. \ref{pc_pics_new} (c)). This final map is only used for robot pose estimation where we want to exclude the effect of the passable obstacles from our calculations. 
Then, each waypoint of the 2D trajectory is vertically projected onto the closest map point to produce the equivalent 3D trajectory. This is achieved using the K-nearest-neighbor search algorithm (KNN)\cite{Knn_kd}. The labelled point cloud map is then used to estimate the roll and pitch angles, denoted by $\gamma$ and $\phi$, respectively, of the two-wheeled robot at each waypoint. The traversability of the 3D path is evaluated by comparing the absolute roll and pitch angles to predetermined threshold values $\gamma_{max}$ and $\phi_{max}$, respectively. If any estimated value exceeds the threshold, the 3D waypoint is regarded as unstable and thus untraversable for the robot. We prevent the 2D planner from building trajectories over untraversable waypoints by setting the corresponding cell and its eight adjacent neighbours as obstacles in the unpassable occupancy grid map.

    

\section{Evaluation}
\vspace{1em}
\label{4}

\begin{figure*}[t!]
\vspace{1em}
\begin{center}
\includegraphics[width=17cm]{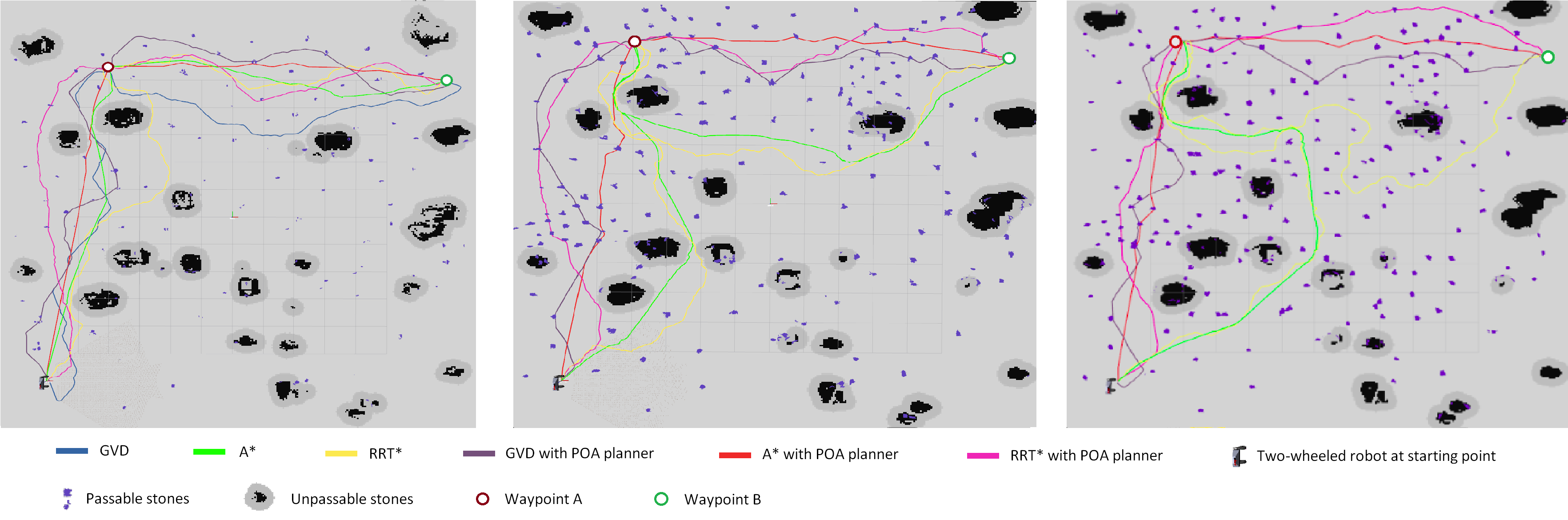}
Setup 1 - 104 passable stones \qquad \qquad  Setup 2 - 159 stones \qquad \qquad Setup 3 - 206 stones
\caption{Three setups for experiments in 2D}
\label{maps}
\end{center}
\vspace{-2em}
\end{figure*}
\vspace{-1em}

\begin{table*}[b!]
\vspace{- 1.5em}
  \begin{center}
    \caption{Experimental results}
    \label{tab:table}
    \begin{tabular}{|l|c c|c c|c c|} 
      \hline
      \multirow{2}{*}{\textbf{Planner name}} & \multicolumn{2}{c|}{\textbf{Setup 1}} & \multicolumn{2}{c|}{\textbf{Setup 2}} & \multicolumn{2}{c|}{\textbf{Setup 3}} \\
      & Distance (m) & traversal time (s) & Distance (m) & traversal time (s) & Distance (m) & traversal time (s) \\
      
      \hline
      GVD & 29.02 & 247 & FAILURE & FAILURE & FAILURE & FAILURE  \\
      A* & 23.54 & \textbf{193} & 30.21 & 257 & 19.5 to A & 148 to A  \\
      RRT* & 27.97 & 251 & 34.92 & 321 & 43.49 & 448  \\
      \hline
      GVD with POA planner & \textbf{26.79} & \textbf{245} & \textbf{26.36} & \textbf{238} & \textbf{26.69} & \textbf{274}  \\
      A* with POA planner & \textbf{22.9} & 204 & \textbf{23.43} & \textbf{240} & \textbf{23.16} & \textbf{220}  \\
      RRT* with POA planner & \textbf{26.01} & \textbf{233} & \textbf{26.53} & \textbf{248} & \textbf{24.91} & \textbf{273}  \\
      \hline
    \end{tabular}
  \end{center}
\end{table*}

\subsection{2D POA path planner}
The idea of this experiment is to test the effect of the POA algorithm on the performance of three standard path planners. We compared POA planner to trajectories generated by $A^\star$, Rapidly Exploring Random Trees$^\star$($RRT^\star$), and GVD. We set $n_{skip}$ to 3 and $n_{clear}$ to 10 for GVD and RRT* POA variants while $n_{skip}$ was 5 and $n_{clear}$ was 20 for A* POA variant. A grid resolution is 0.5 m.


To evaluate the proposed method we developed a simulated environment resembling challenging outdoor conditions. 
For the environment simulation, we used Gazebo simulation engine. ROS\cite{ros} open libraries and Navigation stack were used for the missions execution. Finally, we ran the experiments in a known environment. The labelled grid and point cloud maps are obtained before planning by manually moving the robot in the test environments and collecting the necessary information for the "Point cloud classifier module".



The two-wheeled robot in the simulation is the exact copy of the real prototype presented in Fig. \ref{fig1}. A robot of such size can go over objects up to 28 cm in height and 26 cm in one of the dimensions on the $XY$ plane (\ref{fig1}). These dimensions - are the maximum size of the clearance of our prototype. It means that the requirement for an object to be classified as a passable one is 28 cm in height and 26 cm in width. The length of the obstacle is not limited. Otherwise, it is classified as unpassable. It brings us to the environment for the experiment. There are three setups - three identical maps for the robot to navigate. Fig. \ref{maps} depicts three maps 15 by 15 meters with the identical distribution of unpassable stones - each map contains 22 unpassable objects randomly placed on the maps. The only difference between them is the number of passable stones. Map 1 has 104 randomly distributed passable stones, while Maps 2 and 3 each have 159 and 206 randomly distributed passable stones, respectively. The starting point is in the bottom left corner. The robot has two waypoints to visit. Waypoint $A$ (\ref{maps}) is placed in the top left corner of the map and waypoint $B$ is in the top right corner. The robot's task is simple - build the shortest collision-free path to waypoint A and, after successfully arriving, find the shortest path to the waypoint B and reach it.

We run each standard path planner (GVD, A*, and RRT*) and find a collision-free path from starting point to the waypoint $B$. Then, we integrate into each of these algorithms the pipeline described in section \ref{3}. We run the RRT$^\star$ path planner variants ten times to choose the shortest path by its length (Distance in Table \ref{tab:table}) as one of the metrics of evaluation. The second parameter for comparison is traversal time which is the total travel time from starting point to waypoint $B$. The results of each experiment are presented in Table \ref{tab:table}. The best values are highlighted with bold font inside each experiment for every pair: GVD - GVD with POA planner, A* - A* with POA planner, and RRT* - RRT* with POA planner.

\nointerlineskip
As we can see, in the first setup, the difference between standard algorithms and POA planner is negligible in path length and traversal time. For the second setup (Fig. \ref{maps}, Setup 2), the standard GVD algorithm failed to find a path from the starting point to waypoint $A$. This is because GVD prioritizes the robot's safety by maximizing its clearance from all obstacles so it is prone to failure in densely cluttered environments. Unlike GVD, A* and RRT* coped with the more challenging environment. The path lengths for the standard A* and RRT* increased 28\% and 25\%, respectively, compared to setup 1. Their traversal time also increased by 33\% and 28\%, respectively. On the other hand, All POA variations succeeded to generate paths that excel in their standard variations and show similar results to the first setup. The third setup depicts the densest environment of all three setups (Fig. \ref{maps} Setup 3). The Standard GVD algorithm expectedly failed to find a path in an even more cluttered environment. The Standard A$^\star$ only found a way to waypoint $A$ but failed to build a path to waypoint $B$. On the other hand, The Standard RRT$^\star$ turned out to be the most reliable of the three standard algorithms as it successfully reached the final point in all setups. However, its distance and traversal time drastically increased - by 55\% and 78\% respectively in comparison to the results on the first setup. Thus, we can observe that the POA variations generated paths that don't necessarily depend on the passable obstacles density of the environment. The distance remained almost the same for all three setups and traversal time increased insignificantly. Comparing POA results for the first and the third setups, traversal time increased by 12\% for the GVD algorithm,  by 8\% for the A* algorithm, and 17\% for RRT*. Although, the distances remained the same, the total time to the final point increased as there are more passable stones that slow down the robot each time it passes through them or around them.

\vspace{-0.5em}
\subsection{3D POA path planner}


In many real-life applications, unstructured outdoor environments have uneven terrain. In this experiment, we aim to validate the ability of the POA path planner to generate feasible trajectories on 3D terrain. It allows the two-wheeled robot to find the most optimal routes to complete its mission without the risk of tipping over. We use "Setup 3" map from the previous experiment and make the surface uneven (Fig. \ref{pc_pics_new} (a)). The simulated environment's surface is now three-dimensional, with heights ranging from -1 m to 1 m. We compare the pitch angle and the roll angle at waypoints along the trajectories generated by the 2D POA path planner and the 3D POA path planner.

\begin{figure}[t!]
\begin{center}
\vspace{0.5em}
\subfigure[POA planner 2D and 3D trajectories]{
\resizebox*{4cm}{!}{\includegraphics{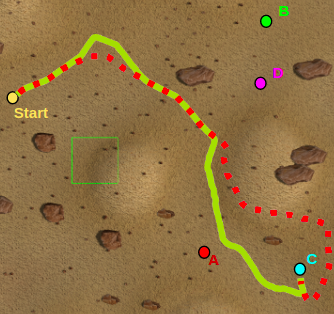}}}
\subfigure[Robot inclination angles following 2D and 3D POA trajectories]{
\resizebox*{4cm}{!}{\includegraphics{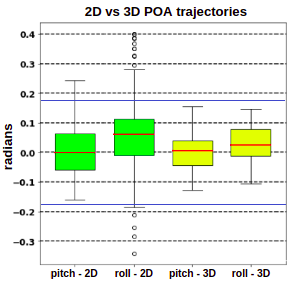}}}
\vspace{-0.5em}
\caption{3D POA path planner experiment setup}
\label{exp_3}
\vspace{-2em}
\end{center}
\end{figure}

Four missions are conducted from the robot's start position to four different goals A, B, C, and D (Fig. \ref{exp_3} (a)). In Fig. \ref{exp_3} (a)) the red dashed line represents a 2D POA path, and the solid green line represents a rebuilt 3D POA path. The maximum pitch and roll angles for the robot to safely follow the paths without losing balance are 0.175 radians. These boundaries are depicted in Fig. \ref{exp_3} (b) with horizontal blue lines. Fig. \ref{exp_3} (b) represents the robot inclination angles while following the paths built by 2D and 3D POA planners going to all four goals - A, B, C, and D. Fig. \ref{exp_3} (a) shows the trajectories from start point to the goal C as an example. As we can see, the 2D POA path builds a trajectory through bumps on the map that makes the robot incline. "pitch - 2D" and "roll - 2D" boxplots in Fig. \ref{exp_3} (b) show that most of the time the trajectory is safe for the robot. However, the maximum pitch and roll angles exceed 0.175 radians boundaries in several waypoints with a maximum of 0.243 radians and 0.4 radians, respectively. It happens since the 2D POA path planner assumes a planar terrain. In such a scenario, the two-wheeled robot will lose stability and fail its mission. 

3D POA planner, on the other hand, shows much better results ("pitch - 3D" and "roll - 3D" boxplots). It deviates much less when following a rebuilt path for a 3D environment and does not breach the thresholds for pitch and roll angles. This experiment demonstrates how critical it is to consider terrain geometry when navigating the two-wheeled robot in unstructured outdoor environments.

In Fig. \ref{exp_3} (b) 3D POA path planner successfully generated feasible trajectories because the defined thresholds are not exceeded at any waypoint along the 3D POA trajectories in all four scenarios - "Start point - goal A", "Start point - goal B", "Start point - goal C", and "Start point - goal D". The maximum pitch and roll angles are 0.156 radians 0.146 radians respectively.
Finally, in the previous experiments, we demonstrated that the 2D POA path planner outperforms standard planners in terms of path length and traversal time in highly cluttered environments. In this experiment, the path lengths of the two path planners' trajectories were also computed to demonstrate that the generated trajectories of the 3D POA path planner and the 2D POA path planner have comparable path lengths as seen in the Table \ref{tab:path_len_exp3}. It is also observed that the 3D POA trajectories avoid dangerous terrain characterized by steep hills.

\begin{table}[h!]
\vspace{- 1em}
    \begin{center}
    \caption{Path length of 2D POA path planner vs 3D POA path planner}
    \resizebox{\columnwidth}{!}{%
    \label{tab:path_len_exp3}
    \begin{tabular}{|l|c c|c c|c c|c c|} 
      \hline
      \multirow{2}{*}{\textbf{metrics}}
      & \multicolumn{2}{c|}{\textbf{Goal A}} & \multicolumn{2}{c|}{\textbf{Goal B}} &\multicolumn{2}{c|}{\textbf{Goal C}} & \multicolumn{2}{c|}{\textbf{Goal D}}\\
      & 2D & 3D & 2D & 3D & 2D & 3D & 2D & 3D\\
      \hline
      Path length(m) & 11.5 & 10.9 & 10  & 10.3 & 15.2 & 14.2 & 10.5  & 10\\
      \hline
    \end{tabular}
    }
    \end{center}
\vspace{- 2.5em}
\end{table}


%
%
%
%
%
%
%
%
%
%
%

\section{Conclusion and Future Work}

This paper presented a new approach to solving path planning problem in highly cluttered areas for two-wheeled robots. The POA solution that we suggest enables a two-wheeled robot to detect, classify, and overcome small (passable) obstacles. Our algorithm was tested in a simulated environment and embedded into A*, GVD, and RRT* path planners. Comparing the performance of the standard planners to their POA variations in highly cluttered environments, the standard path planners struggle to find a path while their POA variations showcase independence from the obstacles and density of the environment. Our method decreases path length and the total travel time to the final destination up to 43\% and 39\%, respectively, comparing to standard path planners such as GVD, A*, and RRT*. The proposed algorithm was extended to work in unstructured 3D environments by taking terrain geometry into account to navigate the two-wheeled robot without the risk of tipping over. The 3D POA path was compared to the planar 2D POA path in the unstructured environment in terms of the pitch angle and roll angle. The 3D POA path planner successfully constructed feasible trajectories for the two-wheeled robot while the 2D POA paths had pitch and roll angles up to 0.243 and 0.4 radians, respectively, which are beyond the robot's stabilization capabilities. These results prove the feasibility of the proposed technology to optimize the navigation of two-wheeled robots in unstructured indoor and outdoor environments.








\bibliographystyle{IEEEtran} 
\bibliography{lit}

\end{document}